# BIOWISH: Biometric Recognition using Wearable Inertial Sensors detecting Heart Activity

Emanuele Maiorana, *Senior Member, IEEE*, Chiara Romano, *Student Member, IEEE*, Emiliano Schena, *Senior Member, IEEE*, and Carlo Massaroni, *Member, IEEE*

*Abstract*—Wearable devices are increasingly used, thanks to the wide set of applications that can be deployed exploiting their ability to monitor physical activity and health-related parameters. Their usage has been recently proposed to perform biometric recognition, leveraging on the uniqueness of the recorded traits to generate discriminative identifiers. Most of the studies conducted on this topic have considered signals derived from cardiac activity, detecting it mainly using electrical measurements thorugh electrocardiography, or optical recordings employing photoplethysmography. In this paper we instead propose a BIOmetric recognition approach using Wearable Inertial Sensors detecting Heart activity (BIOWISH). In more detail, we investigate the feasibility of exploiting mechanical measurements obtained through seismocardiography and gyrocardiography to recognize a person. Several feature extractors and classifiers, including deep learning techniques relying on transfer learning and siamese training, are employed to derive distinctive characteristics from the considered signals, and differentiate between legitimate and impostor subjects. An multi-session database, comprising acquisitions taken from subjects performing different activities, is employed to perform experimental tests simulating a verification system. The obtained results testify that identifiers derived from measurements of chest vibrations, collected by wearable inertial sensors, could be employed to guarantee high recognition performance, even when considering short-time recordings.

## I. INTRODUCTION

Wearable devices are often cited as one of the "next big things" in technological innovation after smartphones. Actually, the global market of wearable technology is rapidly expanding, with a compound annual growth rate (CAGR) of 12% [1]. Earphones, wristbands, and smartwatches are the most popular and widespread among such devices, with headbands, chestbands, goggles, smart clothing, and electronic jewelry also gaining notable attention and relevance. One of the most attractive aspects of this technology is its ability to non-invasively capture physiological data related to the wearer, and the consequent possibility of using the information thus collected for any conceivable application. A very well-know usage of wearable devices is for fitness purposes, with data related to the heart rate or the calorie consumption employed to track the achievement of specific goals, or monitor the health of a subject [2]. A growing interest has recently been associated with the use of wearable devices for medical purposes, due to the opportunity of exploiting the portability of these tools to record critical physiological data over long time periods, while preserving as much as possible the comfort and freedom of movement of the involved subjects [3]. Furthermore, exploiting the feasibility of embedding computing and communication capabilities within them, nowadays wearable devices can be easily integrated within Internet of Things (IoT) frameworks to exchange or share data with other equipment, thus allowing to implement handy value-added services such as smart payment [4]. Especially in reference to the latter possibility, the use of wearable devices has been recently proposed to perform automatic biometric recognition [5]. Actually, the physical, behavioral, or cognitive traits collected by wearable sensors can be effectively processed to derive discriminative features, and exploited to discriminate a legitimate subject wearing the employed devices from potential impostors. A major peculiarity and advantage of recognition systems relying on wearable devices, with respect to traditional approaches implementing desktop solutions, is that the employed biometric traits could be collected autonomously, without asking the involved users to interact with dedicated systems, thus allowing a more convenient and user-friendly acquisition procedure. The recorded data could be then either transmitted to a server where the recognition process has to be carried out, or directly processed within the employed device, in order to decide whether the wearer is really the legitimate owner of the device, before granting logical or physical access to certain areas or services. In addition to a greater ease of use with respect to standard desktop biometric systems, recognition approaches relying on wearable devices could also guarantee improved security, due to the fact that the characteristics recorded by wearable devices typically cannot be captured remotely, making them difficult to steal and replicate, inherently provide liveness detection, and could also be examined to infer about the mental and emotional states of an individual, providing the means to detect compulsion attacks [6]. Moreover, since the acquisition of the interested traits can be done at any time and any place, wearable biometric systems make it feasible to perform continuous recognition, meaning that the identity of a user can be verified throughout the duration of a session, preventing hijacking and avoiding attackers' unauthorized access after an initial successful recognition [7].

Among the traits recordable through wearable devices that have been exploited to perform biometric recognition, the cardiac activity certainly plays a pivotal role. In this context, the present paper evaluates a biometric recognition approach referred to as BIOWISH, since it relies on *W*earable *I*nertial *S*ensors to collect mechanical measurements related to the *H*eart activity. In more detail, we here investigate the feasibility of exploiting relatively new and unexplored modalities for detecting cardiovascular activity, i.e. seismocardiography (SCG) and gyrocardiography (GCG), in order to generate distinctive biometric identifiers. The study here reported actually analyzes:

- the feasibility of recognizing an individual, at an average time distance of 36 days from enrolment, relying on mechanical measurements of the subject's heart activity collected through commercial and inexpensive wearable inertial measurement units (IMUs);
- the effectiveness of deep learning strategies relying on transfer learning and siamese training to generate discriminative representations of the considered SCG and GCG data;
- the dependence of the achievable recognition performance on the placement, over a subject's chest, of the employed wearable inertial sensors;
- the incidence of the time duration of signals collected during the verification process on the achievable equal error rates (EERs);
- the influence of the activity performed by the involved subject on the achievable recognition performance;
- the capability of correctly recognizing the activity that a subject is carrying out, within a set comprising lying down, sitting, standing, and walking, using the considered SCG and GCG measurements.

The paper is organized as follows: Section II provides an overview on the state of the art of wearable biometric recognition systems based on heart activity, including a critical analysis of the studies so far presented on this topic. The techniques considered in the proposed BIOWISH approach to perform biometric recognition based on inertial measurements of the heart activity are presented in Section III. The collected database and the tests carried out on it to evaluate the effectiveness of the proposed verification systems are outlined in Section IV. Conclusions derived from the work done are finally drawn in Section V.

## II. Wearable Biometric Recognition based on Heart Activity

The physiological signals associated with the cardiac muscle functioning have long been studied with the aim of extracting person-specific information, and exploiting it for biometric recognition purposes [8]. In this regard, medical-grade equipment has been traditionally used to detect the small electrical changes resulting from the heart depolarization and repolarization during each cardiac cycle, resorting to multi-lead electrocardiography (ECG). Such approach has made it possible to gain, over time, detailed and reliable knowledge about the person-specific characteristics of the electrical signals characterizing a subject's heart activity [9]. Yet, due to the costs of the used devices, and the commonly uncomfortable acquisition conditions that subjects have to endure, with the need to place multiple electrodes on the chest, wrists, and ankles, these solutions have been typically restricted only to laboratory environments, with limited real-world applications of heart-based biometric recognition so far proposed.

Wearable devices have instead recently allowed to easily collect information related to a subject's cardiac activity even outside controlled environments, thus enabling the design of innovative heart-based recognition solutions that can be conveniently used in daily life. In detail, biometric approaches relying on one- or two-lead ECG data have been proposed by exploiting sensors embedded in chestbands [10], [11], armbands [12], [13], and also electronic textiles, that is, fabrics with embedded electronics [14], [15]. The aforementioned solutions allow to continuously acquire the desired ECG signals without the need for users' intervention, while approaches relying on occasional data recording have been also proposed, as in [16] where commercial wristbands, requiring the involved subjects to touch the device with the opposing hand to collect the interested traits, have been employed. It is yet worth remarking that the quality of ECG signals collected through wearable devices is typically lower than the one achievable when using medical instrumentation [17], with the attainable recognition performance therefore affected by the lower signal-to-noise ratio.

A convenient modality to detect heart activity through wearable devices, recently exploited in several biometric studies, is photoplethysmography (PPG), a non-invasive optical technique relying on pulse oximeters illuminating the skin at close distance with near-infrared (NIR) light [18]. This kind of radiation is absorbed by the blood in vein vessels depending on levels of vasodilation, vasoconstriction, and oxygenation, thus allowing to perceive both variations in blood volume pulse (BVP) due to the cardiac cycle [19], and oxygen saturation (SpO2) in case two distinct wavelengths are used [20]. Measurements can be taken by either sensing transmitted light placing sensors on fingertips [21], or detecting reflected light with sensors placed on the wrist [22]. The collected PPG signals could be either directly processed to extract discriminative features to perform recognition [23], or employed to estimate coarse metrics concerning cardiovascular activity such as heart rate variability (HRV) [24] or metabolic equivalent of task (MET) [25], before exploiting the obtained measures to recognize a subject.

Digital stethoscopes can be also employed to acquire acoustic information on the heart activity through phonocardiography (PCG). Subject-specific characteristics could be actually extracted from the two main sounds of a cardiac cycle, the low-pitch S1 at the closing of mitral and tricuspid valves (systole) and the high-pitch S2 at the closing of aortic and pulmonic valves (diastole) [26]. While quite large medical equipment has been traditionally employed to collect this kind of data, by placing the employed devices over the subjects' chest [27], recent advances in wearable technology, consisting in the development of small sound sensors, could allow improving the acceptability of this acquisition modality, paving the way for practical recognition systems relying on PCG [28].

Besides electrical, optical, and acoustic wearable measurements of heart activity, also mechanical sensing has been exploited to obtain information on a subject's cardiovascular activity. In more detail, SCGs [29] and GCGs [30] allow collecting signals characterizing the low-frequency vibrations caused by heart pumping, when placing sensors on a subject's chest. Both approaches rely on IMUs, implemented using micro-electromechanical sensors (MEMS), to record the desired signal, with three-axis accelerometers employed to acquire SCG traits, and three-axis gyroscopes exploited for GCG data. Discriminative information has been extracted from SCG data considering subjects in sitting [31], [32] and lying [33] conditions, while both SCG and GCG measurements for subjects in standing conditions have been investigated in [34].

The vibrations generated by the cardiac activity can also be detected in body parts far from the heart using ballistocardiography (BCG). Head-mounted devices have been for example employed to extract biometric identifiers from BCG signals in [35] and [36]. However, BCG measurements are notably affected by any contact of the body with external objects, including the floor and the measuring devices, since they may interfere with, or even impede, the body displacement induced by recoil forces [37], with the consequence that this technique commonly provides less informative data than SCG or GCG.

Illustrative examples of the heart-related signals that can be collected through different sensing techniques are given in Figure 1, along with indications about the most relevant events performed during a cardiac cycle, and recognizable from the recorded data. Most of the biometric recognition studies relying on wearable devices to detect heart activity have been performed exploiting electrical and optical measurements through ECG and PPG. Mechanical heart signals have been instead so far underexplored, despite the fact that nowadays SCG and GCG data can be recorded through commercial off-the-shelf devices, with limited dimensions and weights which guarantee high portability and ease of use, affordable costs and long battery life. Actually, SCG and GCG techniques are already spreading rapidly in medical scenarios as alternatives to ECG for monitoring the pre- and post-procedural conditions of patients [39]. In fact, several worthwhile hemodynamic parameters, including pre-ejection periods and systolic time intervals, can be easily and reliably estimated from them [38], [40]. The use of commercial IMUs for heart-based people recognition purposes could therefore represent an interesting opportunity also for the development of innovative biometric systems, with concrete possibilities of implementation in real-world applications, and the present paper represents a thoroughly evaluation of this possibility.

*A. State of the Art Analysis*

In order to verify the effectiveness of the BIOWISH solution here proposed, proper tests are carried out exploiting an in-house multi-session database, specifically collected with the aim of checking the stability over time of the discriminative capabilities of the considered biometric traits. Actually, it is worth noting that the permanence of the employed identifiers is an aspect often neglected in biometric studies, especially when dealing with traits requiring a considerable amount of setup time to be acquired. This happens despite it is a widely known fact that recognition results computed by comparing data from distinct sessions are notably worse than those accomplished on single-session datasets [18], mainly due to aging effects occurring in all biometric traits. It is also worth remarking that, for physiological signals employed as biometric characteristics, the extracted features may be additionally affected by personal habits, such as those related to the current diet, and even by the specific physical condition and emotional state of a subject. Furthermore, when using wearable devices to collect the interested biometric characteristics, also the specific placement of the employed on-body sensors can influence the performed recordings, since it is very unlikely to attach the employed devices each time at the very same position. It is therefore extremely unrealistic to estimate the recognition performance achievable in a real-world application by considering only single-session databases.

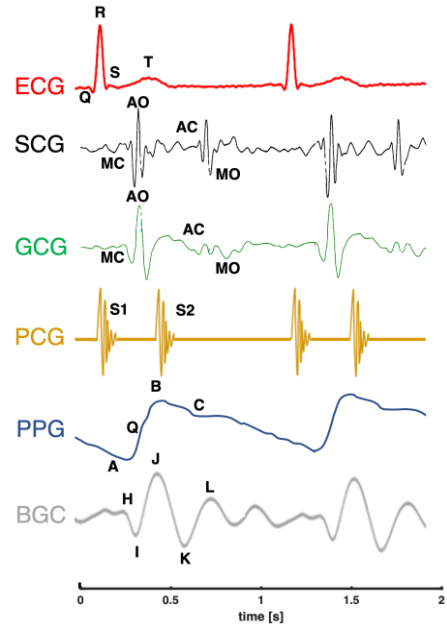

Fig. 1: Simultaneous recordings of ECG, SCG (z-axis), GCG (y-axis), PCG, PPG, and BCG waveforms captured from a healthy subject. Relevant events occurring during a cardiac cycle reported in this picture include the mitral valve closure (MC) and opening (MO), the aortic valve closure (AC) and opening (AO), the S1 sound at the closure of the atrioventricular (mitral and tricuspid) valves, the S2 sound at the closure of the semilunar (aortic and pulmonary) valves, the systolic nadir (A) and end diastole (B). The H wave head-ward deflection begins close to the peak of the R wave and has maximum peak synchronously or near the start of ejection; the I wave foot-ward deflection follows the H wave and occurs early in systole; the J wave largest head-ward wave immediately follows the I wave and occurs late in systole; the K wave foot-ward wave follows the J wave and occurs before the end of systole (adapted from [38]).

When considering heart-based biometric recognition systems relying on wearable devices, only few works have till now investigated the permanence of the exploited identifiers, performing tests on multi-session databases. In more detail, such works are listed in the summary reported in Table I. It can be seen that longitudinal studies have been conducted only on ECG, PPG, and SCG signals, among all the possible heart-based biometric traits. Furthermore, time spans significantly larger than a week between acquisition sessions have been considered only in a handful of studies [13], [14], [21], [22], [41], [42]. Commercial devices have been employed in the majority of studies focused on a single modality, while prototype systems comprising multiple sensors have been used in multi-biometric systems relying on both ECG and PPG. Some works have evaluated the feasibility of recognizing a subject while performing multiple specific activities such as sleeping [42], walking [41], or writing [22], with samples taken in the same conditions compared for recognition purposes. Recordings taken *in the wild*, i.e., with subjects involved in daily routines, have been instead considered only for ECG [10], [14], [15], with recognition there carried out regardless the specific performed activity. The duration of the signals employed to perform recognition is often greater than 10 s, sometimes requiring recordings lasting even more than a minute, especially when coarse metrics such as HRV or MET are derived from the collected data [24], [41].

TABLE I: Summary of state-of-the-art approaches using biometric traits acquired through wearable devices for automatic people recognition. Only studies performed on multi-session datasets are reported.

| Signal | Paper | Subjects | Sessions | Device | Multiple conditions | Time for recognition | Feature | Comparator | Recognition modality | Performance EER | IR |
|---|---|---|---|---|---|---|---|---|---|---|---|
| ECG | [14] | 5 | over 6 months | Commercial | Yes | 6 heartbeats | Wavelet | SVM | CS Identification | - | 70-100% |
|  | [15] | 33 | over 6 weeks | Commercial | Yes | 10 heartbeats | Learned | CNN | CS Identification | - | 95.9% |
|  | [10] | 20 | 6 in 6 days | Commercial | Yes | 3 heartbeats | Statistical | RF | CS Verification | 21.9% | - |
| PPG | [18] | 56 | 2 in 2 days | Prototype | No | 30 hearbeats | Time signal | L2 dist. | OS Verification | 21.5% | - |
|  | [41] | 400 | over 17 months | Commercial | Yes | 5 min | Statistical | OC-SVM | OS Verification | >20% | - |
|  | [22] | 7 | 6 in 50 days | Prototype | Yes | 4 heartbeats | Statistical | RF | CS Identification | - | 90-97% |
|  | [42] | 20 | > 20 days | Commercial | Yes | 10 min | Learned | CNN | CS Identification | - | 55.8% |
|  | [43] | 12 | 4 in 4 days | Prototype | No | 8 s | Learned | CNN | CS Identification | - | 95.7% |
|  | [21] | 100 | 3 in 17 days | Prototype | No | 20 heartbeats | Learned | CNN+RNN | CS Identification | - | 87.1% |
|  | [44] | 17 | 2 in 1 week | Commercial | Yes | 10 s | Spectrogram | CNN | CS Identification | - | 94.9% |
|  | [24] | 74 | over 5 days | Commercial | Yes | 1 min | Statistical | CNN+RNN | CS Verification | 17.9% | - |
| SCG | [31] | 10 | over 1 week | Prototype | No | 1 s | Spectrogram | GMM | CS Verification | 12.0% | - |
| ECG,PPG | [45] | 25 | 2 in two days | Prototype | No | 1 min | Statistical | SVM | CS Identification | - | 92% |
|  | [13] | 53 | 2 in 8 weeks | Prototype | Yes | 1 heartbeat | Time Signal | L2 dist. | OS Verification | 13% | - |

It is also worth observing that, to test the mentioned approaches, most of the performed studies have considered identification scenarios, that is, conditions in which a systems determines the identity of a presented subject exploiting the availability of information taken from multiple users, and system performance is measured in term of identification rate (IR). In more detail, closed-set (CS) identification, in which it is assumed to know all the possible users which could access the system, therefore without the possibility of rejecting an unknown subject, has been always performed in the considered studies. Conversely, verification has been evaluated only in few works, and open-set (OS) conditions have been considered only in a subset of them. For instance, CS verification has been taken into account in [31], estimating for each user a model depending on characteristics of the same impostors also considered for tests. Unfortunately, only OS conditions instead properly resemble operative conditions, where data taken from potential impostors, considered to test a given recognition system, are not available during the enrolment of a legitimate user. The recognition results achieved in these works, normally expressed in terms of equal error rate (EER), are commmonly quite high, always greater than 10%, mainly due to the notable differences between the characteristics of the physiological signals collected in different recording sessions.

Given the exposed state of the art, the aim of the present study is to design a recognition system performing OS verification, able to achieve good recognition performance when comparing heart-related mechanical signals collected by wearable inertial sensors, over short time intervals, and in distinct recording sessions.

### III. Proposed Biometric Recognition System

In order to perform a thoroughly analysis of the discriminative capabilities of SCG and GCG signals, several strategies are employed in the proposed BIOWISH approach to derive, from the considered data, different parametric representations to be used for recognition purposes. In more detail, in the following it is assumed that a generic recording can be expressed through the signals $\{\mathbf{p}_x, \mathbf{p}_y, \mathbf{p}_z\}$, lasting $T$ seconds and sampled at a rate $F$, with $\{x, y, z\}$ being the three axes provided by an accelerometer or gyroscope sensor. The acquired data have to be first preprocessed as described in Section III-A. Then, different parametric representations, to be used when computing the dissimilarity distances between probe and enrolment samples upon which the verification process has to be carried out, are derived as detailed in Section III-B. In the performed tests, the considered feature extractors are trained according to the learning strategies outlined in Section III-C.

#### A. Preprocessing

Given the frequency content of the considered SCG and GCG recordings, the collected data are first low-pass filtered retaining the subband below 25 Hz, and then resampled at a rate $R = 60$ Hz to reduce the computational complexity of the subsequent processing. The resulting signals are then segmented into overlapping frames, using rectangular time windows of length $L = 5$ s with an overlap $O = 80\%$, thus generating a set of $K = \lfloor 1 + \frac{1}{1-O}(\frac{T}{L} - 1) \rfloor$ segments, each comprising $3 \cdot (L \cdot R)$ samples, from an original acquisition. The created frames are treated as individual samples in the considered systems, meaning that a verification process is carried out as a comparison between a query probe, consisting of a single frame lasting $L$ seconds, and an enrolment set, comprising $K$ frames generated from an original recording of duration $T$.

In more detail, a generic frame can be expressed through time-domain coefficients as a $3 \times (L \cdot R) = 3 \times 300$ matrix $\varphi$, where each row of such matrix contains values corresponding to one of the three axes of the employed accelerometer or gyroscope. Alternatively, an expression in the time/frequency domain can be derived as a $25 \times 41 \times 3$ tensor $\Phi$, containing the logarithm of the power spectral densities (PSDs) obtained from the short-time Fourier transforms (STFT) computed over each of the three available axes, using a frequency resolution of 1 Hz over the subband of interest, and 41 Hamming time windows within the considered frame length $L$.

### B. Feature Extractors

A given frame, expressed through the coefficients in $\varphi$ or $\Phi$, can be processed according to multiple approaches in order to derive different discriminative representations. The possibilities evaluated in the performed tests are presented in the following, along with the methods adopted to compare frames acquired during the enrolment and verification phases.

*1) Power Spectral Density and Euclidean Distance:* The simplest representation of a frame can be obtained by averaging the spectra computed through STFT, obtaining a vectorial representation with $25 \cdot 3$ coefficients comprising the frequency contents of all the three considered axes. A verification process can be then performed by comparing a probe frame against all the segments collected during the user's enrolment through an Euclidean distance, and then selecting the lowest dissimilarity measure as representative for the verification probe. If the obtained minimum distance is lower than a selected threshold, the subject is recognized as a legitimate user, otherwise the system considers the presented subject as an impostor.

In this case, it is not required to learn a specific encoding, since simple frequency-domain features are directly exploited to perform recognition. This representation is indicated in the following as $\Phi_{L2}$, given the characteristics of the considered features and of the employed comparator. An analogous approach is employed in [31] to perform biometric recognition exploiting SCG data, with this representation being therefore employed to exemplify a state-of-the-art approach.

*2) Support Vector Machine:* In order to perform an analysis of the frames extracted from an acquisition, a support vector machine (SVM) classifier can be trained to learn an hyperplane separating samples belonging to a specific user from those taken from any other subject. This way, the discriminative characteristics of a user are described through a model, which is employed during verification to compute a similarity score (to be inverted to generate a dissimilarity distance) between a probe frame and the acquisition taken during the user's enrolment. In the performed tests, a radial basis function is employed as SVM kernel. This approach can be effectively applied to the $25 \cdot 41 \cdot 3$ coefficients of the considered short-time spectral template, with the resulting representation being therefore indicated in the following as $\Phi_{SVM}$. The trained strategy employed to learn the sought hyperplane for each enrolled user is outlined in Section III-C.

*3) Convolutional Neural Networks:* The coefficients providing descriptions of the available frames in time or time/frequency domains can be also fed to convolutional neural networks (CNNs) specifically designed with the aim of deriving discriminative representations to be used in a verification process. As already proposed in [34], tensors comprising the time/frequency content of a frame can be used as inputs of networks originally proposed to perform object classification, after being resized in order to comply with the requirements of the considered architectures. The VGG-16 [46] and ResNet18 [47] frameworks are taken into account in the performed tests, since these CNNs have low memory and computational complexity requirements, being therefore plausible to implement them within wearable devices. The obtained representations are reported in the following as $\Phi_{VGG}$ and $\Phi_{RN}$. During the verification process, the features extracted from a probe frame are compared against all the representations generated from the enrolment acquisition of the claimed identity, selecting the lowest computed distance as representative of the comparison process.

With the aim of extracting discriminative information from SCG or GCG frames while further reducing the required computational complexity, tests have been also conducted using other CNNs specifically designed for the considered tasks. In more detail, Table II details a novel architecture, named as $WISHNET_{TF}$, here proposed to extract a discriminative representation $\Phi_{WISHNET}$ with 1024 coefficients from the $25 \times 41 \times 3$ tensor containing the time/frequency content of a frame describing heart activity. The employed architecture comprises 4 convolutional (CONV) layers with rectified linear units (ReLUs) as non-linear activation functions, 3 max-pooling (MP) layers, and 2 dropout (DO) layers.

Features in the time domain are also considered as inputs to another CNN, indicated as $WISHNET_T$ and detailed in Table III, here designed and employed to extract a 128-coefficient representation $\varphi_{WISHNET}$ from the considered $3 \times 300$ matrix. A batch normalization (BN) layer is employed in $WISHNET_T$ to handle the variability of the input data.

It is worth mentioning that additional representations have been investigated in the performed tests, for instance using the time-domain coefficients $\varphi$ as input to L2- and SVM-based comparators, or employing long short-term memory (LSTM) recurrent neural networks (RNNs) with either $\varphi$ or $\Phi$ coefficients as inputs, yet the results obtained in such scenarios are not comparable with those achieved through the representations detailed in this section, and are therefore not considered in the following.

### C. Representation Learning

Different learning strategies have been used to train the SVM- and CNN-based models employed to extract the rep-resentations on the basis of which the verification process is carried out. Specifically, it is assumed that, in addition to the data captured from a specific legitimate user, also a set of acquisitions captured from other subjects is made available during an enrolment process. These additional resources can be exploited to train user-specific binary classifiers, able to discriminate between the current user and potential impostors. Such strategy can be adopted to derive the representations detailed in Section III-B, for methods relying on both SVM and CNN models. Regarding the latter case, when networks originally proposed to perform object classification are exploited, a training based on transfer learning, therefore exploiting the knowledge already acquired from applications in different contexts, can be carried out, in a way similar to what has been exploited in [34] for identification, yet considering a simpler binary classification scenario. Conversely, the networks introduced here have to be trained from scratch, adding to the architectures detailed in Tables II and II a dropout layer and a softmax producing two outputs for the considered binary classification task, and using a cross-entropy (CE) loss function to drive the backpropagation learning algorithm. The resulting representations are therefore indicated in the following as $\Phi_{VGG}^{CE}$, $\Phi_{RN}^{CE}$, $\Phi_{WISHNET}^{CE}$, and $\varphi_{WISHNET}^{CE}$ to remark the usage of a CE loss function during training.

TABLE II: Proposed $WISHNET_{T\_F}$ architecture, used to extract representations $\Phi_{W\_ISHNET}$ from tensors containing the time/frequency coefficients $\Phi$ of a SCG or GCG frame.

| Layer | Filter | Pad | Input | Output |
|---|---|---|---|---|
| Conv | (3×5×3)×128 | [1,2] | 25×41×3 | 25×41×128 |
| ReLu | - | - | 25×41×128 | 25×41×128 |
| MP | 2×2 | - | 25×41×128 | 12×20×128 |
| DO | - | - | 12×20×128 | 12×20×128 |
| Conv | (3×5×128)×256 | [1,2] | 12×20×128 | 12×20×256 |
| ReLu | - | - | 12×20×256 | 12×20×256 |
| MP | 2×2 | - | 6×10×256 | 6×10×256 |
| Conv | (3×5×256)×512 | [1,2] | 6×10×256 | 6×10×512 |
| Relu | - | - | 6×10×512 | 6×10×512 |
| MP | 2×2 | - | 6×10×512 | 3×5×512 |
| DO | - | - | 3×5×512 | 3×5×512 |
| Conv | (3×5×512)×1024 | [0,0] | 3×5×512 | 1×1×1024 |
| ReLu | - | - | 1×1×1024 | 1×1×1024 |

TABLE III: Proposed $WISHNET_T$ architecture, used to extract representations $\varphi_{W\_ISHNET}$ from matrices containing the time-domain coefficients $\varphi$ of a SCG or GCG frame.

| Layer | Filter | Pad | Input | Output |
|---|---|---|---|---|
| Conv | (3×5×1)×32 | [0,0] | 3×300×1 | 1×316×32 |
| BN | - | - | 1×316×32 | 1×316×32 |
| ReLu | - | - | 1×316×32 | 1×316×32 |
| MP | 1×2 | - | 1×316×32 | 1×158×32 |
| DO | - | - | 1×158×32 | 1×158×32 |
| Conv | (1×5×32)×32 | [0,0] | 1×158×32 | 1×154×32 |
| ReLu | - | - | 1×154×32 | 1×154×32 |
| MP | 1×2 | - | 1×154×32 | 1×72×32 |
| DO | - | - | 1×72×32 | 1×72×32 |
| Conv | (1×5×32)×64 | [0,0] | 1×72×32 | 1×68×32 |
| ReLu | - | - | 1×68×32 | 1×68×32 |
| MP | 1×2 | - | 1×68×32 | 1×34×32 |
| Conv | (1×5×64)×64 | [0,0] | 1×34×32 | 1×30×32 |
| ReLu | - | - | 1×30×32 | 1×30×32 |
| MP | 1×2 | - | 1×30×32 | 1×15×32 |
| Conv | (1×5×64)×64 | [0,0] | 1×15×32 | 1×11×64 |
| ReLu | - | - | 1×11×64 | 1×11×64 |
| MP | 1×2 | - | 1×11×64 | 1×5×64 |
| Conv | (1×5×64)×128 | [0,0] | 1×5×64 | 1×1×128 |
| ReLu | - | - | 1×1×128 | 1×1×128 |

However, using the aforementioned learning approach means that a different model has to be trained for each enrolled user. In order to learn a discriminative representation that can be used for any person, a different learning strategy can alternatively be exploited. In more detail, the availability of acquisitions taken from a set of training subjects can be exploited to train networks according to a siamese strategy [48]. Such approach involve the simultaneous update of two parallel networks, with the aim of generating representations having a low distance in case samples from the same subjects are used as inputs, and a high distance in case samples from different subjects are fed to the two networks. Instead of using a softmax with a CE training loss as in the standard classification task, this learning procedure simply computes an Euclidean distance between the representations obtained from the two parallel networks, and computes a contrastive loss function whose minimization is sought through the backpropagation algorithm.

The representations obtained following this learning strategy, indicated in the following as $\Phi_{VGG}^C$, $\Phi_{RN}^C$, $\Phi_{WISHNET}^C$, and $\varphi_{WISHNET}^C$, with the symbol C referring to the contrastive loss function using during training, have the advantage that they can be estimated over a given set of acquisitions, and then employed for any user to be enrolled without the need for training each time a user-specific network. Furthermore, as it will be further detailed in Section IV, the training process can be designed in order to specifically learn representations with the permanence property, by properly selecting the pairs used as inputs for the parallel networks.

The representations considered in the performed tests are listed in Table IV for summary purposes. Regarding the use of SVM, in addition to a strategy relying on the availability of acquisitions from a training set of subjects to learn a binary classifier, also one-class (OC) SVM have been investigated, to evaluate whether effective user-specific classifiers could be estimated exploiting only data taken from the considered user.

Yet, the obtained results have shown that resorting to binary classifiers yields better performance, and therefore OC-SVMs are not considered in the following discussion.

## IV. EXPERIMENTAL TESTS

The database collected to evaluate the proposed BIOWISH recognition systems based on representations derived from SCG and GCG signals is presented in Section IV-A, while the results obtained on it are then reported in Section IV-B.

### A. Employed Database and Settings

Data related to the heart activity of sixteen subjects, twelve male and four female, each without evidence of cardiovascular diseases, have been collected and exploited to evaluate the effectiveness of the proposed approach. In more detail, SCG and GCG signals have been collected, for each involved subject, from five different wearable Xsens Xdot[1] MEMS IMU sensors, each attached at different positions of the subjects' chest using a hypoallergenic adhesive. The landmarks where the sensors have been placed, commonly adopted in medical applications, are shown in Figure 2, and correspond to:

- the auscultation focus of the aortic (A) valve, between the first and second ribs;
- the focus of auscultation of the pulmonary (P) valve, between the first and second rib;
- the mitral (M) valve auscultation focus, between the fourth and fifth ribs;
- the tricuspid (T) valve auscultation focus, between the third and fourth ribs;
- the xiphoid (X) process on the lower part of the sternum.

TABLE IV: Summary of the representations considered in the tests described in Section IV.

| Frame coeff. | Feat. Extractor | Learning | Representation |
|---|---|---|---|
| Time | $WISHNET_T$ | Binary class. | $\varphi^{CE}_{WISHNET}$ |
| | | Siamese | $\varphi^{C}_{WISHNET}$ |
| | Time average | n/a | $\Phi_{L2}$ |
| Time/Frequency | SVM | Binary class. | $\Phi_{SVM}$ |
| | VGG-16, ResNet-18 | TL + bin. cl. | $\Phi^{CE}_{VGG}$, $\Phi^{CE}_{RN}$ |
| | | TL + siamese | $\Phi^{C}_{VGG}$, $\Phi^{C}_{RN}$ |
| | $WISHNET_{TF}$ | Binary class. | $\Phi^{CE}_{WISHNET}$ |
| | | Siamese | $\Phi^{C}_{WISHNET}$ |

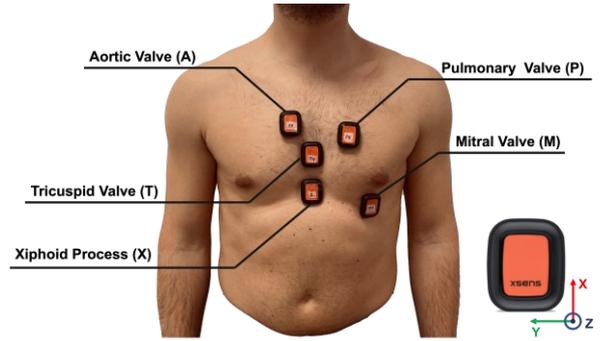

Fig. 2: Anatomical landmarks where the 5 IMU sensors are attached to record free accelerations and angular velocities.

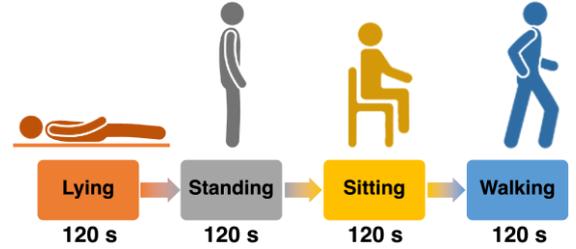

Fig. 3: Considered postures.

Each sensor has a size of 36.30 × 30.35 × 10.80 mm, and a weight of 10.8 g, being therefore extremely light and comfortable to wear, and contains a three-axis accelerometer, a three-axis gyroscope, and a magnetometer. The acceleration signals have been used as SCG data, while the angular velocities as GCG traits. The sensors coordinate system has a body-fixed right-handed Cartesian reference, as shown in Fig. 2. All the IMUs have been synchronized before starting the data acquisition using the companion app, so that all the sensor data were time-synced. The algorithm synchronizes the clocks to an error smaller than 20 μs, which decays to an error smaller than 1.8 ms after 30 min [49]. The acquired data have been stored locally in the memory at a sampling frequency ($F$) of 120 Hz, and then downloaded to the laptop before processed as described in Section III. During a recording session, four different acquisition conditions have been considered for each subject, corresponding to lying, standing, sitting, and walking, as shown in Fig. 3. Each activity has been performed for 120 s while quietly breathing. The existence of permanent characteristics within the considered data has been investigated in the performed tests thanks to the availability of two recording sessions, taken at an average time distance of 36 days, for the involved subjects.

The collected data have been used to learn the representations outlined in Section III. Specifically, since our aim is to test the recognition performance achievable in open-set verification scenarios, the binary classifiers employed to derive the representations $\Phi_{SVM}$, $\varphi^{CE}_{WISHNET}$, $\Phi^{CE}_{VGG}$, $\Phi^{CE}_{RN}$, and $\Phi^{CE}_{WISHNET}$, have been trained using, for each subject, data taken from the first recording session of the considered user as positive class, and data taken from other ten randomly-selected subjects, out of the available ones, as negative class. The recognition rates attainable through these subject-specific models have been estimated considering, for each user, the recordings from the second session as genuine verification probes, and data collected from the other five available subjects, not involved during training, as impostor attempts. The $\Phi^{CE}_{VGG}$ and $\Phi^{CE}_{RN}$ representations have been derived through transfer learning, using as initial CNN weights those defined for an image classification task over Imagenet [50], and then fine-tuning them using a cross-entropy loss function for back-propagation, stochastic gradient descent with momentum (SGDM), a learning rate of 0.0005, and a batch size of 16. The $\Phi^{CE}_{WISHNET}$ and $\varphi^{CE}_{WISHNET}$ representations have been instead obtained training from scratch with SGDM the networks in Table II and III, respectively.

[1]https://www.xsens.com/xsens-dot

The representations relying on siamese training such as $\varphi^{C}_{WISHNET}$, $\Phi^{C}_{VGG}$, $\Phi^{C}_{RN}$, and $\Phi^{C}_{WISHNET}$ have been instead obtained by selecting, for each subject considered as a genuine user, other ten subjects out of the available ones for training purposes, leaving the remaining five for testing. In more detail, the considered networks are trained exploiting only the data from the ten randomly-selected subjects, with pairs whose distance has to be minimized defined by taking two frames extracted from recordings of different sessions of the same subject, and pairs whose distance has to be maximized defined by taking two frames from recordings of different subjects. Following this approach, the networks should explicitly extract permanent characteristics from the considered data, and the learned representations could be thus used also for subjects not involved during the training stage. Network training has been performed using SGDM, a batch size of 64 pairs, and a 0.0001 learning rate.

| Signal | Position | Representation | | | | | | | | | |
|---|---|---|---|---|---|---|---|---|---|---|---|
| | | $\Phi_{L2}$ | $\Phi_{SVM}$ | $\Phi_{VGG}^{CE}$ | $\Phi_{RN}^{CE}$ | $\Phi_{VGG}^{C}$ | $\Phi_{RN}^{C}$ | $\Phi_{WISHNET}^{CE}$ | $\Phi_{WISHNET}^{C}$ | $\varphi_{WISHNET}^{CE}$ | $\varphi_{WISHNET}^{C}$ |
| SCG | Pulmonary | 11.1±0.8 | 8.3±0.7 | 29.5±1.2 | 30.4±1.4 | 24.5±0.9 | 23.5±1.0 | **7.6±0.6** | 9.4±0.7 | 13.6±0.4 | 12.3±0.5 |
| | Aortic | 27.2±0.4 | 27.0±0.5 | 31.2±1.5 | 30.2±1.3 | 28.5±0.9 | 29.4±1.1 | 26.4±0.5 | **18.7±0.4** | 19.2±0.4 | 19.6±0.5 |
| | Xiphoid | 22.9±0.4 | 15.9±0.5 | 29.4±0.9 | 29.1±0.8 | 27.5±0.9 | 27.3±0.8 | 16.0±0.3 | 17.0±0.5 | **12.6±0.4** | 19.2±0.4 |
| | Mitral | 16.1±0.7 | 15.2±0.6 | 29.5±1.1 | 29.2±1.0 | 28.3±0.9 | 27.6±0.8 | 14.6±0.5 | 14.0±0.6 | **10.5±0.4** | 10.7±0.5 |
| | Tricuspid | 23.9±0.8 | 25.6±0.6 | 30.2±1.1 | 30.0±1.3 | 28.6±1.1 | 28.3±0.9 | 27.6±0.6 | 21.5±0.6 | 18.7±0.5 | **16.9±0.4** |
| GCG | Pulmonary | 13.9±0.6 | 14.2±0.7 | 28.6±0.9 | 28.1±0.8 | 18.2±0.9 | 28.1±0.8 | **9.5±0.6** | 10.7±0.7 | 11.7±0.7 | 16.1±0.6 |
| | Aortic | 24.9±0.8 | 29.0±0.9 | 31.2±1.3 | 30.3±1.1 | 30.1±1.1 | 30.6±1.5 | 26.8±0.7 | **17.2±0.5** | 17.7±0.6 | 19.7±0.6 |
| | Xiphoid | 23.1±0.8 | **18.9±0.7** | 29.6±1.0 | 28.2±0.9 | 27.1±0.8 | 27.9±0.9 | 21.7±0.7 | 22.2±0.6 | 19.1±0.5 | 20.4±0.6 |
| | Mitral | 12.4±0.7 | 13.7±0.7 | 27.4±0.8 | 26.5±0.9 | 27.3±0.8 | 28.2±0.9 | 12.4±0.5 | **10.0±0.4** | 18.3±0.6 | 17.6±0.7 |
| | Tricuspid | 18.9±0.8 | 14.6±0.6 | 28.1±0.8 | 27.3±0.9 | 26.7±1.1 | 27.2±0.8 | **9.4±0.4** | 14.7±0.6 | 18.4±0.6 | 22.2±0.7 |

TABLE V: Recognition performance (mean ± standard deviation of EER, in %), obtained considering a single frame lasting 5 s of SCG or GCG data, recorded from subjects in lying posture, as verification probe. Best results for each sensor in bold.

| Signal | Position | Representation | | | | | | | | | |
|---|---|---|---|---|---|---|---|---|---|---|---|
| | | $\Phi_{L2}$ | $\Phi_{SVM}$ | $\Phi_{VGG}^{CE}$ | $\Phi_{RN}^{CE}$ | $\Phi_{VGG}^{C}$ | $\Phi_{RN}^{C}$ | $\Phi_{WISHNET}^{CE}$ | $\Phi_{WISHNET}^{C}$ | $\varphi_{WISHNET}^{CE}$ | $\varphi_{WISHNET}^{C}$ |
| SCG | Pulmonary | 16.7±0.6 | 18.9±0.7 | 28.7±0.9 | 29.1±0.8 | 28.4±0.9 | 27.8±0.8 | 19.1±0.5 | **12.4±0.5** | 19.5±0.7 | 15.8±0.5 |
| | Aortic | 20.3±0.7 | 20.6±0.6 | 29.1±0.9 | 28.7±0.8 | 28.4±1.0 | 27.4±0.9 | 19.4±0.5 | **16.4±1.1** | 18.8±1.4 | 18.3±0.2 |
| | Xiphoid | 23.0±0.8 | 24.9±0.9 | 29.2±0.9 | 28.7±0.8 | 28.3±1.1 | 28.1±1.0 | 26.2±0.5 | 17.9±0.6 | 14.9±0.5 | **14.3±0.4** |
| | Mitral | 22.5±0.8 | 20.5±0.7 | 30.1±0.9 | 31.2±1.2 | 29.8±1.1 | 28.5±1.2 | 19.0±0.7 | 22.8±0.6 | **10.4±0.5** | 13.5±0.6 |
| | Tricuspid | 18.7±0.6 | 17.7±0.5 | 28.9±0.8 | 28.2±0.9 | 27.3±1.0 | 27.1±1.1 | 17.2±0.7 | 14.5±0.6 | 16.7±0.5 | **13.6±0.4** |
| GCG | Pulmonary | 27.2±1.0 | 36.0±1.2 | 31.0±1.2 | 33.4±1.2 | 33.8±1.1 | 34.8±0.9 | 35.0±1.0 | **15.6±1.01** | 36.8±1.4 | 22.1±0.4 |
| | Aortic | 26.3±0.8 | 32.7±1.0 | 33.4±1.1 | 32.4±0.9 | 29.6±1.1 | 29.2±0.9 | 32.7±0.5 | **15.8±1.1** | 20.9±1.4 | 18.5±0.2 |
| | Xiphoid | 23.7±0.8 | 21.1±0.7 | 28.7±0.8 | 29.2±0.7 | 28.6±1.1 | 28.1±1.3 | 21.7±0.5 | **17.2±0.6** | 25.0±1.4 | 17.2±0.8 |
| | Mitral | 16.7±0.7 | 9.8±0.5 | 29.2±0.8 | 29.8±1.3 | 28.1±0.9 | 27.8±0.8 | 12.9±0.5 | **8.7±0.3** | 15.2±0.4 | 11.7±0.4 |
| | Tricuspid | 30.2±1.2 | 37.8±1.4 | 32.4±1.1 | 33.5±1.2 | 34.8±1.1 | 31.3±1.5 | 33.0±1.0 | 22.0±0.7 | 15.2±0.5 | **14.5±0.4** |

TABLE VI: Recognition performance (mean ± standard deviation of EER, in %), obtained considering a single frame lasting 5 s of SCG or GCG data, recorded from subjects in sitting posture, as verification probe. Best results for each sensor in bold.

| Signal | Position | Representation | | | | | | | | | |
|---|---|---|---|---|---|---|---|---|---|---|---|
| | | $\Phi_{L2}$ | $\Phi_{SVM}$ | $\Phi_{VGG}^{CE}$ | $\Phi_{RN}^{CE}$ | $\Phi_{VGG}^{C}$ | $\Phi_{RN}^{C}$ | $\Phi_{WISHNET}^{CE}$ | $\Phi_{WISHNET}^{C}$ | $\varphi_{WISHNET}^{CE}$ | $\varphi_{WISHNET}^{C}$ |
| SCG | Pulmonary | 20.0±0.8 | **10.2±0.4** | 29.1±0.8 | 28.1±0.7 | 28.7±1.1 | 28.1±1.2 | 11.3±0.5 | 12.9±0.6 | 18.8±0.5 | 23.7±0.8 |
| | Aortic | 28.9±0.9 | 19.6±0.8 | 31.2±1.3 | 30.6±1.1 | 29.8±1.0 | 28.7±1.2 | **11.6±0.6** | 21.7±0.6 | 16.9±0.4 | 26.6±0.6 |
| | Xiphoid | 27.2±1.0 | 19.6±0.7 | 30.1±1.3 | 31.2±1.3 | 29.3±1.1 | 28.5±1.0 | 21.2±0.7 | **17.2±0.5** | 24.2±0.7 | 24.0±0.8 |
| | Mitral | 26.1±0.9 | 18.5±0.6 | 29.4±1.3 | 30.2±1.1 | 29.4±1.1 | 30.3±1.2 | 22.0±0.8 | **14.2±0.4** | 17.3±0.5 | 27.0±0.7 |
| | Tricuspid | 27.6±1.0 | **17.5±0.7** | 30.1±1.2 | 31.2±1.3 | 29.4±1.1 | 28.8±1.1 | 20.0±0.7 | 20.6±0.7 | 28.3±0.9 | 26.7±0.8 |
| GCG | Pulmonary | 31.7±1.2 | 21.4±0.8 | 31.0±1.1 | 30.5±1.0 | 29.8±1.0 | 28.8±1.0 | 21.2±0.7 | **14.0±0.4** | 32.8±1.2 | 35.4±1.2 |
| | Aortic | 33.5±1.3 | 22.6±0.7 | 34.5±1.6 | 33.4±1.3 | 32.5±1.1 | 31.8±1.5 | 20.8±0.7 | 21.5±0.9 | **20.5±0.7** | 29.0±0.8 |
| | Xiphoid | 31.4±1.1 | 26.9±0.9 | 33.1±1.1 | 32.5±0.9 | 31.0±1.1 | 30.9±1.5 | 29.0±1.0 | 24.3±0.9 | **21.9±0.7** | 23.7±0.7 |
| | Mitral | 32.1±1.2 | **17.3±0.7** | 31.8±1.3 | 31.0±1.1 | 30.4±1.4 | 30.8±1.5 | 20.0±0.7 | 25.6±1.0 | 19.2±0.8 | 21.7±0.9 |
| | Tricuspid | 29.9±0.9 | 18.5±0.8 | 30.4±1.2 | 32.1±1.3 | 30.1±1.1 | 28.9±1.5 | **17.5±0.6** | 19.9±0.8 | 23.5±0.8 | 30.2±1.2 |

TABLE VII: Recognition performance (mean ± standard deviation of EER, in %), obtained considering a single frame lasting 5 s of SCG or GCG data, recorded from subjects in standing posture, as verification probe. Best results for each sensor in bold.

| Signal | Position | Representation | | | | | | | | | |
|---|---|---|---|---|---|---|---|---|---|---|---|
| | | $\Phi_{L2}$ | $\Phi_{SVM}$ | $\Phi_{VGG}^{CE}$ | $\Phi_{RN}^{CE}$ | $\Phi_{VGG}^{C}$ | $\Phi_{RN}^{C}$ | $\Phi_{WISHNET}^{CE}$ | $\Phi_{WISHNET}^{C}$ | $\varphi_{WISHNET}^{CE}$ | $\varphi_{WISHNET}^{C}$ |
| SCG | Pulmonary | 12.0±0.5 | 11.9±0.5 | 28.6±0.8 | 28.1±0.8 | 26.7±1.1 | 27.2±1.0 | **8.2±0.3** | 9.7±0.4 | 21.5±0.7 | 16.4±0.7 |
| | Aortic | 13.5±0.4 | **10.7±0.5** | 28.9±0.9 | 28.5±1.0 | 26.8±1.1 | 27.3±0.9 | 14.2±0.5 | 13.1±0.5 | 16.3±0.7 | 18.9±0.7 |
| | Xiphoid | 23.6±0.8 | 20.0±0.7 | 28.8±0.9 | 28.6±0.8 | 27.8±1.0 | 27.5±0.9 | 21.0±0.8 | **12.4±0.5** | 18.3±0.7 | 19.4±0.8 |
| | Mitral | 23.7±0.7 | 24.7±0.7 | 29.4±1.0 | 29.3±1.2 | 28.7±1.1 | 29.1±1.4 | 19.1±0.6 | 22.7±1.0 | 25.1±0.9 | **17.0±0.4** |
| | Tricuspid | 21.0±0.6 | 18.8±0.5 | 29.8±0.9 | 28.9±1.2 | 27.8±1.1 | 28.1±1.5 | 17.5±0.6 | 17.8±0.7 | **13.5±0.4** | 17.2±0.6 |
| GCG | Pulmonary | 26.1±0.7 | 22.9±0.6 | 30.1±1.4 | 33.2±1.6 | 30.1±1.1 | 30.7±1.5 | **18.3±0.7** | 21.8±0.8 | 26.0±0.9 | 24.5±0.8 |
| | Aortic | 20.6±0.6 | 18.1±0.5 | 29.9±1.2 | 29.6±1.3 | 28.3±1.1 | 28.0±1.5 | 17.2±0.5 | **16.2±0.5** | 21.5±0.7 | 17.7±0.7 |
| | Xiphoid | 26.1±0.7 | 18.9±0.6 | 30.1±1.3 | 30.0±1.1 | 29.6±0.8 | 29.9±1.0 | 16.7±0.6 | 18.2±0.7 | 18.5±0.7 | **14.0±0.6** |
| | Mitral | 26.9±0.9 | 26.1±0.8 | 32.1±1.4 | 33.5±1.5 | 30.1±1.1 | 30.7±1.7 | 23.9±0.9 | 22.5±0.8 | 23.9±0.8 | **16.7±0.6** |
| | Tricuspid | 17.5±0.6 | 12.5±0.5 | 29.6±0.8 | 30.1±1.1 | 29.1±1.1 | 29.7±1.4 | **11.7±0.5** | 14.0±0.5 | 19.4±0.7 | 13.0±0.6 |

TABLE VIII: Recognition performance (mean ± standard deviation of EER, in %), obtained considering a single frame lasting 5 s of SCG or GCG data, recorded from subjects while walking, as verification probe. Best results for each sensor in bold.

| Signal | Position | Activity | | | |
|---|---|---|---|---|---|
| | | Lying | Sitting | Standing | Walking |
| SCG | Pulmonary | 3.1±0.4 | 10.8±0.5 | 8.2±0.8 | 6.1±0.8 |
| | Aortic | 17.0±0.4 | 11.5±0.5 | 10.8±0.8 | 7.8±0.8 |
| | Xiphoid | 10.1±0.4 | 11.9±0.5 | 13.4±0.8 | 11.8±0.8 |
| | Mitral | 7.6±0.4 | 14.0±0.5 | 12.6±0.8 | 13.8±0.8 |
| | Tricuspid | 19.5±0.4 | 12.0±0.5 | 15.5±0.8 | 10.2±0.8 |
| GCG | Pulmonary | 6.9±0.4 | 28.4±0.5 | 21.9±0.8 | 16.1±0.8 |
| | Aortic | 16.6±0.4 | 18.5±0.5 | 15.7±0.8 | 11.1±0.8 |
| | Xiphoid | 12.6±0.4 | 12.4±0.5 | 18.8±0.8 | 12.4±0.8 |
| | Mitral | 4.8±0.4 | 6.7±0.5 | 10.7±0.8 | 17.9±0.8 |
| | Tricuspid | 12.6±0.4 | 17.8±0.5 | 14.0±0.8 | 7.5±0.8 |
| SCG+GCG | Pulmonary | **2.9±0.4** | 8.5±0.5 | 7.6±0.8 | **4.2±0.8** |
| | Aortic | 15.9±0.4 | 10.5±0.5 | 9.6±0.8 | 5.1±0.8 |
| | Xiphoid | 8.3±0.4 | 9.6±0.5 | 13.1±0.8 | 9.8±0.8 |
| | Mitral | 3.9±0.4 | **4.5±0.5** | **7.2±0.8** | 11.4±0.8 |
| | Tricuspid | 9.2±0.4 | 12.0±0.5 | 12.6±0.8 | 6.47±0.8 |

TABLE IX: Verification performance, in terms of EER (mean ± standard deviation, in %), obtained considering a single frame lasting 5 s of SCG or GCG data as verification probe, and jointly using multiple representations. Best results for each activity in bold.

### B. Experimental Results

The recognition results obtained when considering SCG or GCG signals made of a single frame, therefore lasting $L = 5$ s, as verification probe, are reported in terms of EERs in Tables V-VIII, with the performance obtained for different activities reported in distinct tables. The results are referred to comparisons of recordings captured during different acquisition sessions, yet considering the same activity and the same sensor location in both enrolment and verification stages. All the representations outlined in Section III are taken into account. Cross-validation tests have been conducted changing, at each iteration, the sets of 10 subjects employed to learn the adopted representations. Furthermore, for each considered activity and subject, $T = 90$ s from the first recording session have been randomly extracted from the available recordings and employed as enrolment data.

First of all, it can be observed that permanent characteristics could be actually derived from SCG and GCG signals of limited duration, and exploited to recognize a subject during acquisition sessions carried out some time after an enrolment. SCG data commonly allow achieving better performance than GCG ones, however these latter are more discriminative for subjects in sitting conditions, testifying the need to consider both kinds of data to better handle multiple scenarios. The obtained results also show that methods relying on transfer learning, exploited for single-session identification tests in [34], cannot even achieve performance attained with simple approaches such as those based on Euclidean distances. On the other hand, the networks here proposed in Tables II and III commonly guarantee the best recognition rates among the considered strategies. In more detail, representations obtained from coefficients in both the time ($\varphi$) and time/frequency ($\Phi$) domains can be effectively exploited to perform biometric verification, with better results obtained with different sets depending on the considered physiological signal and considered sensor position. Overall, it can be noticed that training the employed networks with a siamese strategy is preferable on most occasions, with respect to the usage of binary cross-entropy learning, especially for signals collected while sitting. The activity during which the best recognition rates are achieved turns out to be that in which the subjects are lying, while the best sensor positions are the ones located at the left side of the chest, that is, at the pulmonary and the mitral valves.

In order to improve the achievable recognition rates, tests have been done by combining, through a sum operator, the scores obtained with the four most discriminative representations, that is, $\Phi_{WISHNET}^{CE}$, $\Phi_{WISHNET}^{C}$, $\varphi_{WISHNET}^{CE}$, and $\varphi_{WISHNET}^{C}$. The results achieved when using SCG and GCG signals, both individually and jointly by fusing the scores obtained with each trait, are reported in Table IX. In can be observed that, exploiting a multi-biometric approach, low EERs can be achieved for all the considered activities. Figure 4 then shows the recognition performance achievable for increasing durations of the verification probes. The reported EERs have been obtained taking the average of the scores obtained for consecutive frames, and considering signals collected by a sensor placed at the pulmonary valve, which is the location guaranteeing the overall best recognition rates. The obtained results demonstrate that EERs less than 5% can be achieved when exploiting recordings lasting 20 s and collected while carrying out all the considered activities, with very low error rates obtained especially for subjects in lying conditions. The walking scenario is easier to be handled than the sitting and standing ones, most probably due to the fact that the used IMUs are able to retain characteristics related to the subject behavior, in addition to information associated to the cardiac activity. Recognition capabilities in sitting and standing conditions are similar, with the latter one being slightly more favorable using sensors at the pulmonary valve. The use of IMUs to capture heart-induced vibrations on a subject's chest is therefore an effective solution which could be exploited to recognize a user even several days after the initial enrolment.

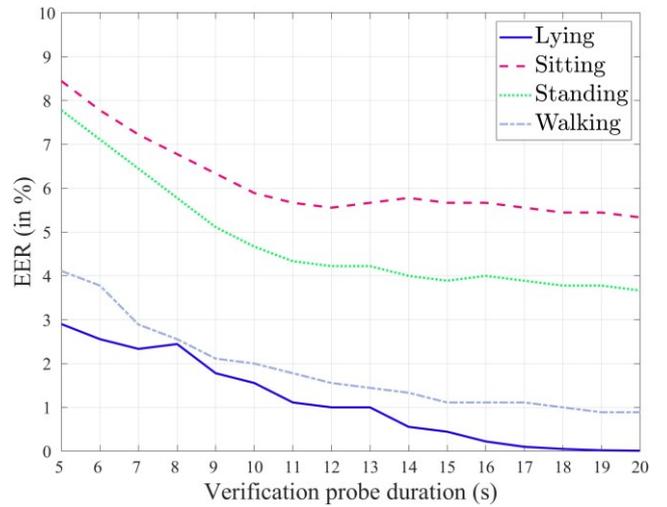

Fig. 4: Mean EERs (in %) achievable in different conditions for verification probes with increasing temporal duration, jointly using SCG and GCG data collected by a sensor positioned at the pulmonary valve.

| Signal | Position | $\Phi_{SVM}$ | $\Phi_{VGG}^{CE}$ | Representation $\Phi_{RN}^{CE}$ | $\Phi_{WISHNET}^{CE}$ | $\varphi_{WISHNET}^{CE}$ |
|---|---|---|---|---|---|---|
| SCG | Pulmonary | 85.8±0.4 | **92.1±0.5** | 86.7±0.8 | 89.7±0.8 | 90.2±0.8 |
|  | Aortic | 86.8±0.4 | 89.2±0.5 | **93.4±0.8** | 89.2±0.8 | 83.7±0.8 |
|  | Xiphoid | 88.8±0.4 | **93.9±0.5** | 92.0±0.8 | 93.8±0.8 | 86.5±0.8 |
|  | Mitral | **94.7±0.4** | 91.5±0.5 | 93.0±0.8 | 93.3±0.8 | 93.1±0.8 |
|  | Tricuspid | 87.7±0.4 | 88.4±0.5 | **91.6±0.8** | 89.0±0.8 | 89.7±0.8 |
| GCG | Pulmonary | **86.9±0.4** | 82.3±0.5 | 74.5±0.8 | 85.5±0.8 | 70.0±0.8 |
|  | Aortic | 86.0±0.4 | 86.8±0.5 | **89.1±0.8** | 88.2±0.8 | 67.2±0.8 |
|  | Xiphoid | 82.2±0.4 | 80.0±0.5 | 80.0±0.8 | **82.7±0.8** | 78.9±0.8 |
|  | Mitral | 68.8±0.4 | 77.1±0.5 | 74.7±0.8 | 82.9±0.8 | **83.1±0.8** |
|  | Tricuspid | 80.5±0.4 | 80.7±0.5 | 81.1±0.8 | **88.8±0.8** | 82.4±0.8 |

TABLE X: Overall accuracy (mean ± standard deviation, in %) of the activity recognition task, obtained considering a single frame lasting 5 s of SCG or GCG data as recognition probe. Best results for each position in bold.

As previously mentioned, the results so far presented have been obtained when comparing signals captured from subjects performing the same activity during both enrolment and verification. However, as already observed in [34], it would be desirable to have the ability to recognize a person without having to know *a priori* the performed activity. Unfortunately, as shown in [34], the signals collected during distinct activities are significantly different, making it hard to recognize a person using SCG or GCG collected in conditions different than the ones carried out during the enrolment. In order to handle such differences, in [34] it has been suggested to train classifiers using data collected while performing all the admissible activities. Although such approach is feasible, following this strategy implies a significant deterioration of the achievable recognition performance, with respect to scenarios focused on a specific activity, as shown in [34]. In order to preserve the capability of attaining high recognition rates, it is here proposed to perform a different approach relying on a two-stage user recognition process, during which the activity a user is performing has to be first determined, and then a verification model specifically learned for that condition is employed. In more detail, such activity recognition task is here addressed as a standard classification problem, with the same representations outlined in Section III-B employed to extract characteristics peculiar for each considered condition, and a cross-entropy function employed as loss to be minimized when training the considered neural networks.

The obtained results, expressed in terms of overall accuracy, are shown in Table X for both SCG and GCG data, and sensors placed at different locations. For reasons of compactness, the results obtained with the least performing representation, that is, $\Phi_{L2}$, are not reported. Cross-validations tests have been carried out by randomly selecting, at each of 5 iterations, 10 subjects out of the available ones to collect the data upon which the training process is carried out, with samples taken from the remaining 6 subjects used as testing probes in a 4-activity classification process. From the obtained results it is possible to notice that, in the activity recognition task, the representations obtained through transfer learning as in [34] are highly effective, differently from what has been observed

| Signal | Position | | | | |
|---|---|---|---|---|---|
|  | Pulmonary | Aortic | Xiphoid | Mitral | Tricuspid |
| SCG | 92.5±0.4 | 94.5±0.5 | 93.4±0.8 | 95.3±0.8 | 90.4±0.8 |
| GCG | 88.3±0.4 | 91.5±0.5 | 83.5±0.8 | 83.2±0.8 | 90.3±0.8 |
| SCG+GCG | **96.7±0.4** | **97.4±0.5** | **93.8±0.8** | **96.7±0.8** | **91.4±0.8** |

TABLE XI: Overall accuracy (mean ± standard deviation, in %) of the activity recognition task, obtained considering a single frame lasting 5 s of SCG or GCG data as recognition probe, and jointly using multiple representations. Best results for each position in bold.

for the user verification task, especially for SCG data.

Yet, it has to be observed that, in the proposed two-stage user recognition process, errors in the recognition of the performed activity could imply severe degradation of the overall verification performance. In order to improve the attainable results, it could be thus beneficial to combine the information extracted from the representations derived from the available signals. Table XI reports the classification accuracies achievable when combining the scores obtained using the four most effective representations, that is, $\Phi_{VGG}^{CE}$, $\Phi_{RN}^{CE}$, $\Phi_{WISHNET}^{CE}$, and $\varphi_{WISHNET}^{CE}$. The reported results show that very accurate activity recognition can be accomplished for sensors placed at all the considered positions, especially when combining the information derived from both SCG and GCG data.

The recognition capabilities could be further enhanced by considering recognition probes longer than 5 s. Specifically, Figure 5 shows the activity classification accuracy achievable by taking decisions relying on the mean of the scores computed for consecutive frames, considering signals recorded at the pulmonary valve, that is, the position guaranteeing the best verification results as shown in Table IX. The reported plots show that the combined use of the recorded SCG and GCG data allows to perfectly recognize the performed activity, among the considered ones, when exploiting recognition probes lasting approximately 20 s. The confusion matrix reported in Figure 6 provides more insights regarding the most difficult activities to recognize. Specifically, it can be seen that, when using a single frame of SCG data, significant errors are made in correspondence of acquisitions taken in two similar conditions such as sitting and standing, while the lying and walking activities can be recognized very effectively. The combined use of SCG and GCG data notably improves the ability to recognize the carried out activity, with the possibility to discern also sitting and standing conditions when exploiting signals lasting 20 s. Given the availability of SCG and GCG recordings of proper length, the proposed two-stage user recognition approach would therefore allow to determine the identity of a person guaranteeing the verification performance shown in Figure 4, without the need to know *a priori* the performed activity within a set of predefined ones.

## V. CONCLUSIONS

In this paper, a biometric recognition approach relying on wearable inertial sensors to detect mechanical heart activity, namely BIOWISH, has been proposed and tested. Several representations have been considered to extract discriminative information from the considered SCG and GCG signals, both exploiting transfer learning from existing networks, and introducing novel networks processing data in either time or time/frequency domains. The proposed systems have been tested considering recordings collected through commercial and inexpensive wearable IMUs from subjects carrying out four different activities, and placing sensors at different positions to evaluate the configuration guaranteeing the best verification performance.

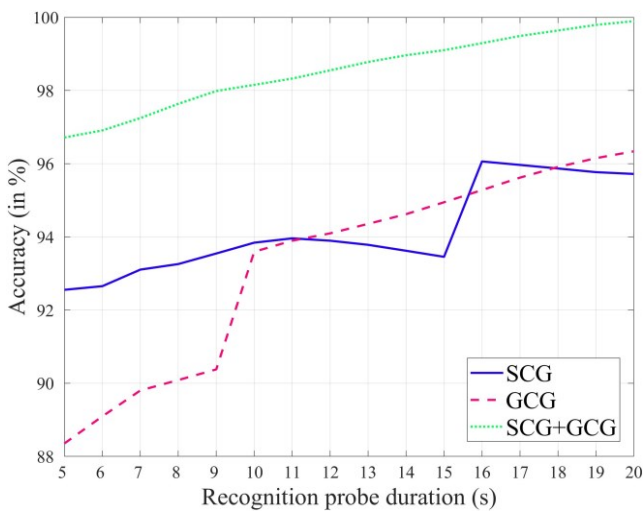

Fig. 5: Overall mean accuracy (in %) of the activity recognition task, for recognition probes with increasing temporal duration, using SCG and GCG data collected by a sensor positioned at the pulmonary valve.

Fig. 6: Confusion matrix of the activity recognition task, obtained considering a sensor placed at the pulmonary valve, and using as recognition probe either a single frame lasting 5 s of SCG data (in blue), a frame with SCG and GCG data (in red), or SCG and GCG recordings lasting 20 s (in green).

It has been observed that the networks here proposed are able to achieve the best verification performance among the considered representations, especially when a siamese strategy is adopted for training. The joint use of both SCG and GCG data, and the exploitation of recordings lasting approximately 20 s, allow to reach rather high recognition performance, especially for subjects in lying conditions and data acquired through sensors placed in correspondence to the pulmonary valve. A two-stage process, requiring to determine the carried out activity before extracting person-specific characteristics, has been also proposed to perform automatic people recognition without having to know *a priori* the performed activity. In this regard, networks trained with transfer learning have been shown to be particularly effective in extracting activity-dependent characteristics. The joint usage of SCG and GCG data proved to be essential to achieve high performance in the activity recognition task, a fundamental prerequisite to perform an accurate user verification process. The obtained results testify that heart-induced vibrations, measured through IMU wearable devices placed on a subject's chest, could represent a reliable biometric identifier to be exploited in automatic verification systems.

REFERENCES

[1] A. Ometov *et al.*, "A survey on wearable technology: History, state-of-the-art and current challenges," *Computer Networks*, vol. 193, 2021.
[2] J. Chaki, N. Dey, F. Shi, and R. S. Sherratt, "Pattern mining approaches used in sensor-based biometric recognition: a review," *IEEE Sensors Journal*, vol. 19, no. 10, pp. 3569–3580, 2019.
[3] A. Mosenia, S. Sur-Kolay, A. Raghunathan, and N. Jha, "Wearable medical sensor-based system design: A survey," *IEEE Transactions on Multi-Scale Computing Systems*, vol. 3, pp. 124–138, 2017.
[4] V.-H. Lee, J.-J. Hew, L.-Y. Leong, G. W.-H. Tan, and K.-B. Ooi, "Wearable payment: A deep learning-based dual-stage sem-ann analysis," *Expert Systems with Applications*, vol. 157, p. 113477, 2020.
[5] J. Blasco, T. M. Chen, J. Tapiador, and P. Peris-Lopez, "A survey of wearable biometric recognition systems," *ACM Computing Surveys (CSUR)*, vol. 49, no. 3, pp. 1–35, 2016.
[6] A. Sundararajan, A. Sarwat, and A. Pons, "A survey on modality characteristics, performance evaluation metrics, and security for traditional and wearable biometric systems," *ACM Computing Surveys*, vol. 52, no. 2, pp. 1–36, 2019.
[7] S. Khan, S. Parkinson, L. Grant, N. Liu, and S. McGuire, "Biometric systems utilising health data from wearable devices: Applications and future challenges in computer security," *ACM Computing Surveys*, vol. 53, no. 4, pp. 1–29, 2020.
[8] F. Agrafioti, J. Gao, and D. Hatzinakos, "Heart biometrics: Theory, methods and applications," in *Biometrics*, J. Yang, Ed. InTech, 2011.
[9] A. Rathore, Z. Li, W. Zhu, Z. Jin, and Z. Jin, "A survey on heart biometrics," *ACM Computing Surveys*, vol. 53, no. 6, 2020.
[10] F. Lehmann and D. Buschek, "Heartbeats in the wild: A field study exploring ECG biometrics in everyday life," in *CHI SIGCHI*, 2020.
[11] S. Sprager, R. Trobec, and M. Juric, "Feasibility of biometric authentication using wearable ecg body sensor based on higher-order statistics," in *MIPRO*, 2017.
[12] J. Blasco and P. Peris-Lopez, "On the feasibility of low-cost wearable sensors for multi-modal biometric verification," *Sensors*, vol. 18, no. 9, p. 2782, 2018.
[13] M. Martinho, A. Fred, and H. Plácido da Silva, "Towards continuous user recognition by exploring physiological multimodality: An electrocardiogram (ECG) and blood volume pulse (BVP) approach," in *ISSI*, 2018.
[14] C. Ye, B. Vijaya Kumar, and M. Tavares Coimbra, "Human identification based on ecg signals from wearable health monitoring devices," in *ISABEL*, 2011.
[15] B. Pourbabaee, M. Howe-Patterson, E. Reiher, and F. Benard, "Deep convolutional neural network for ECG-based human identification," in *CMBES*, 2018.
[16] V. Chandrashekhar, P. Singh, M. Paralkar, and O. Tonguz, "PulseId: The case for robustness of ECG as a biometric identifier," in *IEEE MLSP*, 2020.
[17] F. Fu, W. X. Y. An, B. Liu, X. Chen, S. Zhu, and J. Li, "Comparison of machine learning algorithms for the quality assessment of wearable ECG signals via lenovo H3 devices," *Journal of Medical and Biological Engineering*, vol. 41, no. 2, pp. 231–240, 2021.
[18] J. Sancho, Á. Alesanco, and J. García, "Biometric authentication using

the ppg: a long-term feasibility study," *Sensors*, vol. 18, no. 5, p. 1525, 2018.

[19] U. Yadav, S. Abbas, and D. Hatzinakos, "Evaluation of PPG biometrics for authentication in different states," in *IEEE ICB*, 2021.

[20] R. Donida Labati, V. Piuri, F. Rundo, F. Scotti, and C. Spampinato, "Biometric recognition of PPG cardiac signals using transformed spectrogram images," in *IAPR TC4 Workshop on Mobile and Wearable Biometrics*, 2020.

[21] D. Hwang, B. Taha, D. Lee, and D. Hatzinakos, "Evaluation of the time stability and uniqueness in PPG-based biometric system," *IEEE TIFS*, vol. 16, 2021.

[22] Y. Cao, Q. Zhang, F. Li, S. Yang, and Y. Wang, "PPGPass: Nonintrusive and secure mobile two-factor authentication via wearables," in *INFOCOM*, 2020.

[23] R. Donida Labati, V. Piuri, F. Rundo, and F. Scotti, "Photoplethysmographic biometrics: A comprehensive survey," *Pattern Recognition Letters*, vol. 156, pp. 119–125, 2022.

[24] D. Ekiz, Y.S.Can, Y. Dardagan, F. Aydar, R. Kose, and C. Ersoy, "End-to-end deep multi-modal physiological authentication with smartbands," *IEEE Sensors Journal*, vol. 21, no. 13, pp. 14 977–14 986, 2021.

[25] S. Vhaduri, S. Dibbo, and W. Cheung, "HIAuth: A hierarchical implicit authentication system for IoT wearables using multiple biometrics," *Access*, 2021.

[26] M. Bugdol and A. Mitas, "Multimodal biometric system combining ECG and sound signals," *Pattern Recognition Letters*, vol. 38, pp. 107–112, 2014.

[27] A. Spadaccini and F. Beritelli, "Performance evaluation of heart sounds biometric systems on an open dataset," in *IEEE DSP*, 2013.

[28] X. Cheng, P. Wang, and C. She, "Biometric identification method for heart sound based on multimodal multiscale dispersion entropy," *Entropy*, vol. 22, no. 238, pp. 1–21, 2020.

[29] D. Rai, H. Thakkar, S. Rajput, J. Santamaria, C. Bhatt, and F. Roca, "A comprehensive review on seismocardiogram: Current advancements on acquisition, annotation, and applications," *Mathematics*, vol. 9, no. 2243, 2021.

[30] T. J. , "Gyrocardiography: A new non-invasive monitoring method for the assessment of cardiac mechanics and the estimation of hemodynamic variables," *Scientific Reports*, vol. 7, no. 6823, 2017.

[31] E. Vural, S. Simske, and S. Schuckers, "Verification of individuals from accelerometer measures of cardiac chest movements," in *IEEE BIOSIG*, 2013.

[32] A. Bui, Z. Yu, and F. Bui, "A biometric modality based on the seismocardiogram (SCG)," in *International Conference and Workshop on Computing and Communication (IEMCON)*, 2015.

[33] P.-Y. Hsu, P.-H. Hsu, and H.-L. Liu, "Exploring seismocardiogram biometrics with wavelet transform," in *IEEE International Conference on Pattern Recognition (ICPR)*, 2020.

[34] E. Maiorana and C. Massaroni, "Biometric recognition based on heart-induced chest vibrations," in *IEEE IWBF*, 2021.

[35] J. Hernandez, D. McDuff, and R. Picard, "Bioinsights: Extracting personal data from "still" wearable motion sensors," in *IEEE International Conference on Wearable and Implantable Body Sensor Networks (BSN)*, 2015.

[36] J. Hebert, B. Lewis, H. Cai, K. Venkatasubramanian, M. Provost, and K. Charlebois, "Ballistocardiogram-based authentication using convolutional neural networks," in *arXiv*, 2018.

[37] O. I. , "Ballistocardiography and seismocardiography: A review of recent advances," *IEEE Journal of Biomedical and Health Informatics*, vol. 19, no. 4, pp. 1414–1427, 2015.

[38] S. Sieciński, P. S. Kostka, and E. J. Tkacz, "Gyrocardiography: A review of the definition, history, waveform description, and applications," *Sensors*, vol. 20, no. 22, p. 6675, 2020.

[39] O. I. , "Novel wearable seismocardiography and machine learning algorithms can assess clinical status of heart failure patients," *Circulation: Heart Failure*, vol. 11, no. 1, 2018.

[40] M. J. Tadi, E. Lehtonen, T. Hurnanen, J. Koskinen, J. Eriksson, M. Pänkäälä, M. Teräs, and T. Koivisto, "A real-time approach for heart rate monitoring using a hilbert transform in seismocardiograms," *Physiological measurement*, vol. 37, no. 11, p. 1885, 2016.

[41] S. Vhaduri and C. Poellabauer, "Multi-modal biometric-based implicit authentication of wearable device users," *IEEE Transactions on Information Forensics and Security*, vol. 14, no. 12, pp. 3116–3125, 2019.

[42] G. Retsinas, P. Filntisis, N. Efthymiou, E. Theodosis, A. Zlatintsi, and P. Maragos, "Person identification using deep convolutional neural networks on short-term signals from wearable sensors," in *ICASSP*, 2020.

[43] E. Lee, A. Ho, Y. Wang, C. Huang, and C. Lee, "Cross-domain adaptation for biometric identification using photoplethysmogram," in *IEEE ICASSP*, 2020.

[44] E. Piciucco, E. Di Lascio, E. Maiorana, S. Santini, and P. Campisi, "Biometric recognition using wearable devices in real-life settings," *Pattern Recognition Letters*, vol. 146, pp. 260–266, 2021.

[45] A. Diaz Alonso, C. Travieso, and M. D. J.B. Alonso, "Biometric personal identification system using biomedical sensors," in *IEEE CCIS*, 2016.

[46] K. Simonyan and A. Zisserman, "Very deep convolutional networks for large-scale image recognition," *arXiv preprint arXiv:1409.1556*, 2014.

[47] H. Kaiming, X. Zhang, S. Ren, and J. Sun, "Deep residual learning for image recognition," in *IEEE conference on computer vision and pattern recognition (CVPR)*, 2016.

[48] G. Koch, R. Zemel, and R. Salakhutdinov, "Siamese neural networks for one-shot image recognition," in *ICML deep learning workshop*, 2015.

[49] E. D'Alcala', J. Voerman, J. Konrath, and A. Vydhyanathan, "Xsens dot wearable sensor platform white paper," *White Paper*, 2021.

[50] O. Russakovsky, J. Deng, H. Su, J. Krause, S. Satheesh, S. Ma, Z. Huang, A. Karpathy, A. Khosla, M. Bernstein *et al.*, "Imagenet large scale visual recognition challenge," *Int. journal of computer vision*, vol. 115, no. 3, pp. 211–252, 2015.

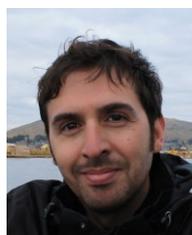

**Emanuele Maiorana** (IEEE SM) received the Ph.D. degree in biomedical, electromagnetism, and telecommunication engineering with European Doctorate Label in 2009. He is currently Assistant Professor at Roma Tre University. His research interests are in the area of digital signal and image processing, with specific emphasis on biometric recognition. He is an Associate Editor of the IEEE Transactions on Information Forensics and Security, and has served as General Chair of the 9th IEEE International Workshop on Biometrics and Forensics in 2021.

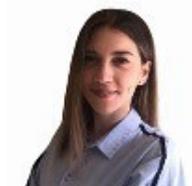

**Chiara Romano** (IEEE Student Member) received the M.Sc. degree (cum laude) in Biomedical Engineering from the Universita' Campus Bio-Medico di Roma (UCBM) in 2021, where she is currently Ph.D. student. Her research interests include the development and test of wearable devices based on IMU sensors for medical and real world applications.

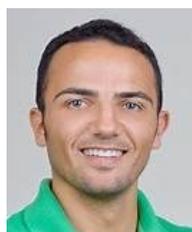

**Emiliano Schena** (IEEE SM) received the Ph.D. degree in 2007. He is Full Professor of Measurements with UCBM. His main research interests include the design and assessment of wearable systems for vital signs monitoring, applications of fiber optic sensors in medicine, and thermal ablation for cancer removal. He became the Chair of the Italy Chapter of the IEEE Sensors Council in 2018.

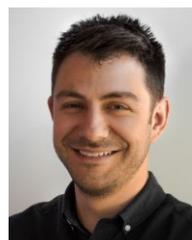

**Carlo Massaroni** (IEEE SM) received the Ph.D. degree in biomedical engineering in 2017. He is currently Assistant Professor of Measurements and Biomedical Instrumentation with UCBM. His research interests include the design, development, and test of wearable devices and unobtrusive measuring systems for medical applications. He serves as Chair of the Wearable Sensors TC of the Italy Chapter of the IEEE Sensors Council since 2020 and he is Associate Member of the TC on Wearable Biomedical Sensors and Systems of the IEEE EMBS.